\documentclass[letterpaper, 10 pt, conference]{ieeeconf}
\IEEEoverridecommandlockouts
\usepackage{cite}
\usepackage{amsmath,amssymb,amsfonts}
\usepackage{algorithmic}
\usepackage{graphicx}
\usepackage{textcomp}
\usepackage{xcolor}
\usepackage{booktabs}
\usepackage{import}
\usepackage{bm}
\usepackage{siunitx}
\usepackage{gensymb}
\usepackage{comment}
\usepackage[font=footnotesize]{caption}
\usepackage[font=footnotesize]{subcaption}
\usepackage{arydshln}
\usepackage{tabularx}

\usepackage[letterpaper, top=21.7mm, bottom=15.7mm, left=17.4mm, right=17.4mm]{geometry}

\usepackage{tikz}
\usetikzlibrary{external}
\usepackage{pgfplots}
\pgfplotsset{compat=1.9}
\usepgfplotslibrary{statistics}
\usepgfplotslibrary{fillbetween}

\usepackage{titlesec}
\titlespacing{\section}{0pt}{*1.3}{*1.3}

\newcommand{\R}{\mathbb{R}}

\makeatletter
\let\NAT@parse\undefined
\makeatother
\usepackage{hyperref}

\begin{document}



\title{\LARGE \bf Visibility Maximization Controller for Robotic Manipulation}


\author{\authorblockN{Kerry He\authorrefmark{1}, Rhys Newbury\authorrefmark{1}, Tin Tran\authorrefmark{1}, Jesse Haviland\authorrefmark{2}, Ben Burgess-Limerick\authorrefmark{2}, Dana Kuli\'c\authorrefmark{1}, Peter Corke\authorrefmark{2} \\and Akansel Cosgun\authorrefmark{1}
\thanks{\authorrefmark{1}Monash University, Australia}%
\thanks{\authorrefmark{2}Queensland University of Technology Centre for Robotics, Australia}%
}}

\maketitle
\begin{abstract}

Occlusions caused by a robot's own body is a common problem for closed-loop control methods employed in eye-to-hand camera setups. We propose an optimization-based reactive controller that minimizes self-occlusions while achieving a desired goal pose. The approach allows coordinated control between the robot's base, arm and head by encoding the line-of-sight visibility to the target as a soft constraint along with other task-related constraints, and solving for feasible joint and base velocities. The generalizability of the approach is demonstrated in simulated and real-world experiments, on robots with fixed or mobile bases, with moving or fixed objects, and multiple objects. The experiments revealed a trade-off between occlusion rates and other task metrics. While a planning-based baseline achieved lower occlusion rates than the proposed controller, it came at the expense of highly inefficient paths and a significant drop in the task success. On the other hand, the proposed controller is shown to improve visibility to the line target object(s) without sacrificing too much from the task success and efficiency. Videos and code can be found at: \href{http://rhys-newbury.github.io/projects/vmc/}{rhys-newbury.github.io/projects/vmc/}.

\end{abstract}


\section{Introduction}

Traditionally, visual sensing and robotic manipulation are combined in an open-loop fashion by first ``looking" and then ``moving"~\cite{corke1994high}. Open-loop control is popular due to its simplicity, especially for fixed-base, high-accuracy manipulators operating in static environments~\cite{lenz2015deep,mahler2017dex}. On the other hand, closed-loop control methods such as visual servoing aim to increase the accuracy of the manipulation task by utilizing continuous sensor feedback, and are also robust to dynamically changing environments~\cite{hutchinson1996tutorial}. Closed-loop vision-based controllers depend on continuous feedback of the target's position. Therefore, losing visibility of the target is detrimental to the controller performance, which can be caused either by the target being outside the camera's field-of-view (FoV) or if another object occludes the line-of-sight (LoS) between the camera and the target.

\begin{figure}[]
\centering
\resizebox{1.0225\linewidth}{!}{\input{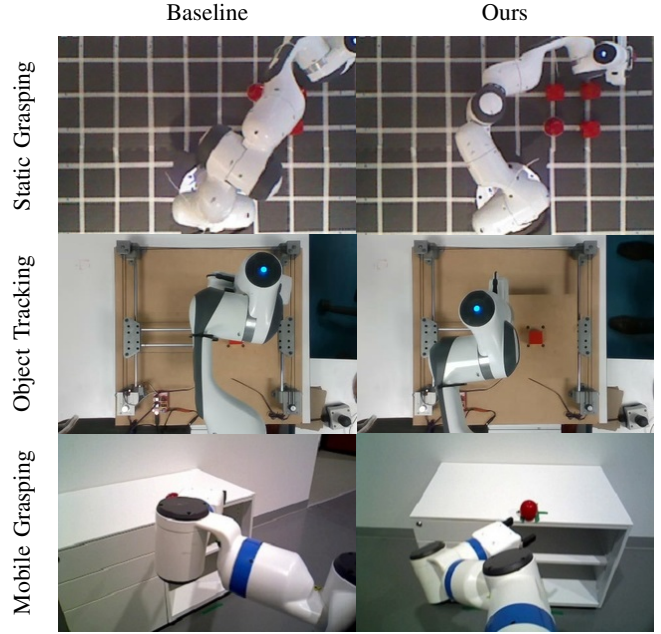}}
\caption{Scenarios comparing the proposed visibility maximization controller (right) and standard baseline controllers (left) for top-down grasping in a cluttered environment (top), top-down grasping of a moving object (middle), and grasping with a mobile manipulator (bottom). In each scenario, the aim is to either grasp or reach towards a red object.}
\vspace{-0.15cm}
\label{fig:intro_fig}
\end{figure}

One class of occlusions is \textit{self-occlusions}, which we define to be LoS occlusions caused by the robot's own linkages. Examples of self-occlusions are shown in the left column of Fig.~\ref{fig:intro_fig}. Most state-of-the-art approaches to closed-loop manipulation do not explicitly address the self-occlusion problem. A common way to avoid the self-occlusion problem is to use eye-in-hand camera setups which enables an unobstructed view from the end-effector. However, eye-in-hand setups suffer from minimum sensing distance limitations of depth cameras~\cite{haviland2020control} or may not be suitable in situations where global information of the workspace is required~\cite{flandin2000eye}. In these cases, fixed eye-to-hand camera setups, or cameras mounted onto a link relatively far from the end-effector, such as a robotic head, are preferred. 

Using the redundant degrees of freedom (DoF) of a robotic manipulator, it is possible to perform the task of controlling the manipulator to a target while simultaneously avoiding self-occlusions. For complex robotic systems which include a mobile base or a camera mounted onto an independently controllable end-effector, additional redundant DoFs can be exploited to improve controller performance. 

In this paper, we present an optimization-based reactive controller, which aims to maximize the visibility to target(s) by minimizing self-occlusions, while simultaneously reaching a goal configuration. Our work extends the framework of~\cite{haviland2021holistic}. An objective function is constructed for a robot to perform a reaching task that aims to keep the target in sight and avoid self-occlusions, while keeping the robot away from singularities and joint limits. The controller then solves for the joint velocities that minimize this objective function at each time-step using a quadratic program (QP).

The contributions of this paper are as follows:
\begin{itemize}
    \item An optimization-based task-space controller to avoid self-occlusions. Our approach generalizes to stationary robot arms as well as mobile manipulators, stationary and controllable camera configurations, and stationary or moving objects. The proposed approach achieves coordinated control between a mobile base, robotic arm, and camera, including branched controlled camera configurations. 
    \item Simulated and real-world experiments that validate the performance of the proposed algorithm.
\end{itemize}






\section{Related Work}\label{sec:related-work}

\subsection{Task-Space Control}

Task-space (or operational-space) control formulates the control objective in the task-space of the robot. For redundant robotic manipulators, it is possible to perform multiple tasks simultaneously using the redundant DoF, such as avoiding collisions with obstacles~\cite{sciavicco1988solution} or maximizing manipulability~\cite{haviland2020purelyreactive}, while simultaneously reaching towards a goal. 

There are two main ways to perform task-space control in the literature: stack-of-tasks or an optimization-based approach. The stack-of-tasks approach produces an analytic solution which projects secondary tasks into the null space of a primary task~\cite{sciavicco1988solution}. The optimization-based approach relaxes the strict hierarchy of tasks, and instead combines all tasks into a single optimization problem where each task can be weighted independently~\cite{haviland2020neo}. The stack-of-tasks approach guarantees a strict order of priority for tasks, however, it is difficult to directly enforce inequality constraints, such as collision avoidance~\cite{mansard2009unified}. A potential field method is a common way to model inequality constraints for stack-of-tasks approaches, however, it does not guarantee that the inequality constraints are satisfied, and can lead to intractable problems in complex environments~\cite{khatib1986real}. 

Our work uses an optimization-based approach to directly incorporate self-occlusion inequality constraints, and to balance competing objectives to achieve desirable behaviors for different scenarios.

\subsection{Occlusion Avoidance}

In papers that address the occlusion avoidance problem, the most common problem definition is to avoid occlusions caused by other stationary or moving objects blocking an object of interest from the camera~\cite{baumann2010path,marchand1998dynamic,nicolis2018occlusion}. FoV constraints are also commonly incorporated into visual servoing controllers, either implicitly through controlling the camera towards the target~\cite{marchand1998dynamic}, or explicitly by modeling the FoV boundaries as constraints~\cite{baumann2010path}. Another common class of problems are when the object itself partially occludes its own feature points from the camera~\cite{tarabanis1996computing}. 



A less explored problem class is self-occlusions, which occur when the robotic linkages block the LoS to an object. Compared to other occlusion types, this problem is unique because, in addition to controlling the camera to avoid occluded viewpoints, the robot link causing the occlusions can also be controlled. There are a few previous works that consider the self-occlusion problem. In~\cite{sugiura2007real}, self-occlusion avoidance is implemented as an application of self-collision avoidance. A stack-of-tasks approach is used where self-collisions are modeled using a potential field method. Similarly, \cite{suzuki2020coordinated} also uses a stack-of-tasks controller, but instead projects the gradient of an occlusion-avoiding objective function into the null space of higher-order tasks. In \cite{logothetis2018model}, a model predictive controller (MPC) is presented which can explicitly impose a minimum distance between the robotic arm and LoS in task-space. However, an MPC typically requires a pre-planned trajectory, and is formulated as a complex optimization problem which is computationally demanding particularly for high DoF robotic systems~\cite{grandia2019feedback}. FoV criteria are incorporated in \cite{suzuki2020coordinated} and \cite{logothetis2018model} in image-space and task-space respectively to ensure the target object remains in frame of the camera.

Our work differs from these previous works in a few ways. Firstly, we achieve self-occlusion avoidance by using a weighted optimization task-space controller, which as opposed to stack-of-tasks approaches allows the controller to smoothly transition between motion towards the target and self-occlusion avoidance, and is able to directly incorporate self-occlusion avoidance inequality constraints. Secondly, prior self-occlusion avoidance works are restricted to a single stationary target object. Our approach is able to account for both multiple objects and moving objects. Thirdly, we apply our approach to different robotic platforms, and empirically measure the benefits that self-occlusion avoidance brings for closed-loop visual servoing.






\section{Robot Model}\label{sec:model}

\begin{figure}[b]
\vspace{-0.4cm}
\centering{
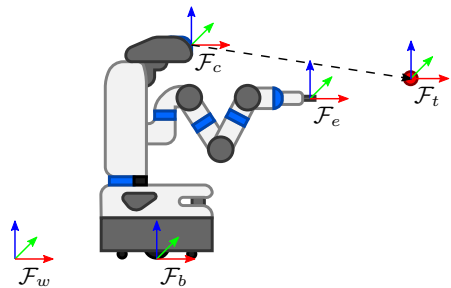
\caption{The mobile manipulator is divided into four key frames, including the base frame $\mathcal{F}_b$, end-effector frame $\mathcal{F}_e$, camera frame $\mathcal{F}_c$, and target frame $\mathcal{F}_t$. The dashed ray models the LoS of the camera. Red, green, and blue represent the $x$, $y$, and $z$ axes of each frame respectively.}
\label{fig:fetch_frames}
}
\end{figure}

We model the robot as a mobile manipulator with a controllable eye-to-hand camera mounted onto the mobile base. This framework can easily generalize to simpler cases by removing DoFs of the system. Important reference frames include the world frame $\mathcal{F}_w$, the mobile base frame $\mathcal{F}_b$, the robotic manipulator end-effector frame $\mathcal{F}_e$, the camera head frame $\mathcal{F}_c$, and the target frame $\mathcal{F}_t$, as depicted in Fig.~\ref{fig:fetch_frames}. 

In the following, we denote ${}^aT_b \in \textrm{SE}(3)$, ${}^aR_b \in \textrm{SO}(3)$, and $\bm{t}^a_b \in \R^3$ as the transformation, rotation, and translation from frame $\mathcal{F}_a$ to frame $\mathcal{F}_b$ respectively. The unit vectors ${}^a\bm{x} \in \R^3$, ${}^a\bm{y} \in \R^3$, and ${}^a\bm{z} \in \R^3$ represent the basis vectors corresponding to the coordinates of frame $\mathcal{F}_a$.

The pose of the mobile base $^wT_b$ is described by its coordinates $(x_b, y_b)$ and rotation $\theta_b$. To describe the kinematics of the base, a virtual base frame $\mathcal{F}_{b'}$ is introduced to model infinitesimal rotations $\delta_\theta$ and translations $\delta_x, \delta_y$ about the base~\cite{haviland2021holistic}. The infinitesimal transforms allow for the base to be included in the holistic differential kinematics of the mobile manipulator. The forward kinematics of the mobile base can be expressed as
\begin{equation}
    {}^wT_{b'} = {}^wT_{b}(x_b, y_b, \theta_b)~{}^bT_{b'}(\bm{q}_b),
\end{equation}
where $\bm{q}_b \in \R^\beta$ and ${\lVert \bm{q}_b \rVert}\rightarrow0$, where $\beta$ is the number of degrees of freedom for the robotic base. For a non-holonomic base which can only translate forwards along the $^b\bm{x}$-axis $\bm{q}_b=\left[\delta_\theta, \delta_x\right]^\top \in \R^2$. 

The forward kinematics of the end-effector and camera can subsequently be derived relative to the base frame, and expressed as
\begin{subequations}
    \begin{align}
        &{}^wT_{e} = {}^wT_{b'}(x_b, y_b, \theta_b, \bm{q}_b)~{}^{b'}T_{e}(\bm{q}_e, \bm{q}_c)\\
        &{}^wT_{c} = {}^wT_{b'}(x_b, y_b, \theta_b, \bm{q}_b)~{}^{b'}T_{c}(\bm{q}_e, \bm{q}_c),
    \end{align}
    \label{eqn:fkine}%
\end{subequations}
where $\bm{q}_e \in \R^n$ and $\bm{q}_c \in \R^m$ represent the joint variables of the robotic arm and camera head respectively, and $n$ and $m$ are the number robotic arm joints and camera head joints respectively. These equations account for branched controlled camera configurations where the manipulator and camera head joints may be coupled. Using the approach in~\cite{haviland2020systematic}, the differential kinematics of the end-effector and camera are described by
\begin{subequations}
    \begin{align}
        &\bm{\nu}_{e} = J_{e}(x_b, y_b, \theta_b, \bm{q})~\dot{\bm{q}}\\
        &\bm{\nu}_{c} = J_{c}(x_b, y_b, \theta_b, \bm{q})~\dot{\bm{q}},
    \end{align}
    \label{eqn:diff-kin}
\end{subequations}
where $\bm{q} = \left[ \bm{q}_b, \bm{q}_e, \bm{q}_c \right]^\top$ are all of the joints of the robot, $\bm{\nu}_{(\cdot)}=\left[v_x, v_y, v_z, \omega_x, \omega_y, \omega_z\right]^\top \in \R^6$ is a 6D twist, and $J_{e} \in \R^{6\times N}$ and $J_{c} \in \R^{6\times N}$ represents the Jacobian of the end-effector and the camera respectively, where $N=\beta+n+m$. These Jacobians can be decomposed into their linear $J^v$ and angular $J^\omega$ components, such that $J=\left[J^v, J^\omega \right]^\top$.

\section{Visibility Maximizing Controller}~\label{sec:controller}
The proposed reactive Visibility Maximizing Controller (VMC) is an extension of the holistic controller presented in~\cite{haviland2021holistic}. It is expressed as the following QP
\begin{subequations}
    \begin{align}
        \min_{\dot{\bm{q}}, \bm{\epsilon}} \quad & \frac{1}{2}(\dot{\bm{q}}^\top \Lambda_q \dot{\bm{q}} + \bm{\epsilon}^\top \Lambda_\epsilon \bm{\epsilon}) - {(J^m_e)}^\top\dot{\bm{q}}_e - \theta\delta_\theta \label{eqn:qp-obj}\\
        \textrm{subj. to} \quad  & J_e\dot{\bm{q}} - \bm{\epsilon}_e= \bm{\nu}_e \label{eqn:qp-ee}\\
                                 & J^\omega_c\dot{\bm{q}} - \bm{\epsilon}_c= \bm{\omega}_c  \label{eqn:qp-cam}\\
                                 & v_l(\dot{\bm{q}}) \leq \xi \frac{\mu_l - \mu_s}{\mu_i - \mu_s}, \quad l=1, \ldots, n \label{eqn:col}\\
                                 & \dot{d}_l(\dot{\bm{q}}) - \epsilon_v \leq \zeta \frac{d_l - d_s}{d_i - d_s}, \quad l=1, \ldots, n \label{eqn:qp-viscol}\\
                                 & \dot{q}_l \leq \eta \frac{\rho_l - \rho_s}{\rho_i - \rho_s}, \quad l = \beta+1, \ldots, N \label{eqn:qp-qlim}\\
                                 & \dot{\bm{q}}^- \leq \dot{\bm{q}} \leq \dot{\bm{q}}^+ \label{eqn:qp-qdlim}\\
                                 & \epsilon_v \geq 0,
    \end{align}
    \label{eqn:qp}%
\end{subequations}
where $\bm{\epsilon} = \left[\bm{\epsilon}_e^\top, \bm{\epsilon}_e^\top,\epsilon_v \right]^\top$ are slack variables used to model soft constraints, and $\Lambda_{(\cdot)}$ are positive diagonal weight matrices described in more detail in Sec.~\ref{sec:obj-weight}. 

From~\cite{haviland2021holistic}, VMC incorporates joint velocity minimization, manipulability maximization, and base orientation objectives~\eqref{eqn:qp-obj}, where $J^m_e$ is the manipulability matrix of the robotic arm described in~\cite{haviland2020purelyreactive} and $\theta$ is the angle between the base orientation and the end-effector. An end-effector velocity control constraint~\eqref{eqn:qp-ee} moves the end-effector towards a target destination, where $\bm{\nu}_e$ is the twist which drives the end-effector in a straight line towards the goal. Joint limit constraints~\eqref{eqn:qp-qlim} formulated as a velocity damper where $\eta \in \R_+$ adjusts the aggressiveness of the velocity damper, $\rho_l$ is the distance from the nearest joint limit, $\rho_i$ is the influence distance at which the velocity damper becomes active, and $\rho_s$ is the minimum distance allowed from the joint limit, and joint velocity limit constraints~\eqref{eqn:qp-qdlim} where $\dot{\bm{q}}^-$ and $\dot{\bm{q}}^+$ are the minimum and maximum joint limits respectively, are also adopted. From~\cite{haviland2020neo}, VMC also incorporates obstacle avoidance constraints~\eqref{eqn:col}, where $\xi \in \R_+$ adjusts the aggressiveness of the velocity damper, $\mu_l$ is the distance between the obstacle and robotic arm, $\mu_i$ is the influence distance, $\mu_s$ is minimum distance allowed by the constraint, and $v_l(\dot{\bm{q}})$ is the rate of change of $\mu_l$.

To achieve the desired occlusion avoidance behavior, the following novel additions are made to the QP. Sec.~\ref{sec:fov} introduces the FoV occlusion avoidance constraint~\eqref{eqn:qp-cam} and Sec.~\ref{sec:occlusion-avoid} introduces the self-occlusion avoidance constraints~\eqref{eqn:qp-viscol}. Furthermore, a method to relax the end-effector angular velocity equality constraint corresponding to the last three rows of~\eqref{eqn:qp-ee} into an inequality constraint is presented in Sec.~\ref{sec:vel-control} to allow for more flexible motion of the robot.

\subsection{FoV Occlusion Avoidance} \label{sec:fov}

When the camera is mounted onto a controllable end-effector or head, it should be controlled to always point towards the target so that it remains within the camera's FoV. This criteria can be specified as the following desired angular velocity of the camera
\begin{equation}
    \bm{\omega}_c = \chi~\Phi\left({}^c\bm{x}, \bm{t}^c_t\right),
\end{equation}
where $\chi \in \R_+$ is a positive gain value, $\bm{t}^c_t$ is the vector from the camera frame to the target frame, and $\Phi\colon\R^3\times\R^3\to\R^3$ is a function which finds the shortest combination of rotations about the $x$, $y$, and $z$-axes from the first vector to the second.

The method used to control the end-effector in~\cite{haviland2021holistic} is used to control the camera to achieve this angular velocity. This is done by introducing the soft equality constraint
\begin{equation}
        J^\omega_c \dot{\bm{q}} - \bm{\epsilon}_c = \bm{\omega}_c,
\end{equation}
where $\bm{\epsilon}_c \in \R^3$ are slack variables for the camera angular velocity, while simultaneously minimizing the magnitude of joint velocities in the objective function. Coordinated control between the camera and the rest of the robot is achieved by incorporating camera control and end-effector control constraints into the same QP.

\subsection{Self-Occlusion Avoidance} \label{sec:occlusion-avoid}


To prevent the robotic manipulator from intruding into the camera's FoV and occluding the target, the LoS between the camera and object is treated as a collision. The collision avoidance method between an obstacle and the robot from~\cite{haviland2020neo} is extended for this application. By treating the LoS similarly to an additional robotic linkage, the avoiding the LoS is analogous to a self-collision avoidance problem.

\begin{figure}[t]
\centering{
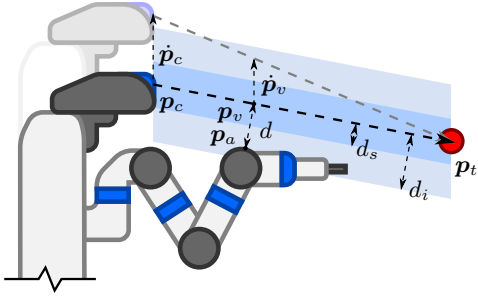
\caption{Camera occlusion avoidance as modeled as a collision avoidance problem. The distance $d$ between the LoS and each linkage is constrained to be greater than $d_s$ using a velocity damper activated within $d_i$. The velocity of the LoS critical point $\dot{\bm{p}}_v$ is related to the velocity of the camera $\dot{\bm{p}}_c$ through similar triangles.}
\vspace{-0.35cm}
\label{fig:vision}
}
\end{figure}

To model the self-occlusion avoidance problem, first consider one of the robotic linkages. We define the critical points $\bm{p}_a \in \R^3$ and $\bm{p}_v \in \R^3$ as the points on the robotic linkage and LoS that are closest to each other. There will be a different pair of critical points for each linkage of the robot. The distance between the critical points is
\begin{equation}
    d = {\lVert \bm{p}_a - \bm{p}_v \rVert},
\end{equation}
and the rate of change of the distance can be expressed as
\begin{equation}
    \dot{d} = \hat{\bm{n}}^\top_{av}(\dot{\bm{p}}_a - \dot{\bm{p}}_v),\label{eqn:dist-roc}
\end{equation}
where $\hat{\bm{n}}^\top_{av}$ is the unit vector pointing from $\bm{p}_a$ to $\bm{p}_v$. As $\bm{p}_a$ is a point on the robotic manipulator, its velocity can be expressed using the differential kinematics~\eqref{eqn:diff-kin}
\begin{equation}
    \dot{\bm{p}}_a = J^v_a \dot{\bm{q}},\label{eqn:arm-col-fkine}
\end{equation}
where $J^v_a$ is the Jacobian relating the linear velocity of the point $\bm{p}_a$ on the robotic manipulator to the robot's joint velocities.

The LoS can be treated as a linkage extending between the camera frame $\mathcal{F}_c$ and the target frame $\mathcal{F}_t$. The velocity $\dot{\bm{p}}_v$ can be found using superpositioned similar triangles shown in Fig.~\ref{fig:vision} as
\begin{equation}
    \dot{\bm{p}}_v = \kappa J^v_c \dot{\bm{q}} + (1-\kappa)\dot{\bm{p}}_t, \quad \kappa=\frac{{\lVert \bm{p}_v - \bm{p}_t \rVert}}{{\lVert \bm{p}_c - \bm{p}_t \rVert}}.\label{eqn:vision-col-fkine}
\end{equation}
From~\eqref{eqn:arm-col-fkine} and~\eqref{eqn:vision-col-fkine}, the rate of change $\dot{d}$ can be expressed as a function of the joint velocities.

A velocity damper~\cite{faverjon1987local} is used to impose a constraint on $\dot{d}$ such that the distance $d$ between two collision objects never drops below a stopping distance $d_s$, by enforcing a limit on $\dot{d}$ which approaches zero as $d$ approaches $d_s$. This takes the form
\begin{equation}
    \dot{d}_l \leq \zeta \frac{d_l - d_s}{d_i - d_s}, \quad l=1, \ldots, n, \label{eqn:vel-damper}
\end{equation}
where $d_i$ is the influence distance at which the velocity damper becomes active, and $\zeta \in \R_+$ adjusts the aggressiveness of the velocity damper. A separate constraint exists for each linkage. Substituting in~\eqref{eqn:dist-roc},~\eqref{eqn:arm-col-fkine} and~\eqref{eqn:vision-col-fkine} gives
\begin{equation}
        \hat{\bm{n}}^\top_{av} \left(J^v_a - \kappa J^v_c
        \right) \dot{\bm{q}} - \epsilon_v \leq \zeta \frac{d - d_s}{d_i - d_s} + (1-\kappa)\hat{\bm{n}}^\top_{av}\dot{\bm{p}}_t,
\end{equation}
for each linkage. This inequality is formulated as a soft constraint using the slack variable $\epsilon_v \in \R_+$, as the end-effector will eventually need to enter the LoS to the object when grasping it.

This method is easily generalizable to situations where occlusion avoidance is desired for multiple objects by modeling multiple LoS to each object. Introducing a new slack variable for each additional object encourages the controller to avoid occluding as many obstacles as possible.

\subsection{End-Effector Orientation} \label{sec:vel-control}

The end-effector objective~\eqref{eqn:qp-ee} is originally defined in~\cite{haviland2021holistic} to move to a specified end-effector orientation. However, in situations where end-effector orientation is not important for the success of the task, the end-effector angular velocity constraints can be removed, which frees up the robotic manipulator's DoF to give it a better chance to perform occlusion avoidance. This can be particularly advantageous for static-base robotic manipulators and static cameras which typically have fewer DoF. However, it can still be advantageous to constrain the orientations of the end-effector. For example, for top-down grasping, the end-effector must point downwards to avoid collisions with the table. 

\begin{figure}[t]
\centering{
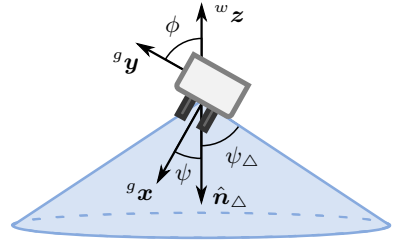
\caption{Diagram showing the cone region defined by the half-apex angle $\psi_\triangle$ and normal vector $\hat{\bm{n}}_\triangle$ that the cone is projected in the direction of. The direction the gripper points in ${}^gx$ is constrained to be within this region.}
\vspace{-0.35cm}
\label{fig:cone}
}
\end{figure}

This constraint is modeled as restricting the forward-facing gripper axis ${}^g\bm{x}$ within a specified cone defined by its half-apex angle $\psi_\triangle$ and normal vector $\hat{\bm{n}}_\triangle$, as shown in Fig.~\ref{fig:cone}. The angle $\psi$ from $\hat{\bm{n}}_\triangle$ to ${}^g\bm{x}$, and the axis $\hat{\bm{n}}_\psi$ of this angle are found as


\begin{equation}
    \psi = \textrm{acos}\left(  \hat{\bm{n}}_\triangle^\top {}^g\bm{x}  \right), \qquad
    \hat{\bm{n}}_\psi = \hat{\bm{n}}_\triangle \times {}^g\bm{x}.
\end{equation}

The velocity of $\psi$ can be related to the joint velocities as
\begin{align}
    \dot{\psi} &= \hat{\bm{n}}_\psi^\top \bm{\omega}_e = \hat{\bm{n}}_\psi^\top J^\omega_e \dot{\bm{q}}.
\end{align}
Subsequently, a velocity damper of the form \eqref{eqn:vel-damper} can be formed to constrain $\psi$ to be within the specified cone.
\begin{equation}
    \hat{\bm{n}}_\psi^\top J^\omega_e \dot{\bm{q}} \leq \frac{\textrm{cos}(\psi) - \textrm{cos}(\psi_\triangle)}{1 - \textrm{cos}(\psi_\triangle)} \label{eqn:cone}
\end{equation}

In addition to constraining the direction the gripper points in, for tasks such as top-down grasping, it is also advantageous to ensure the ${}^g\bm{y}$ axis is parallel to the grasping surface (the world $xy$ plane). This is equivalent to enforcing a perpendicularity constraint between ${}^g\bm{y}$ and ${}^w\bm{z}$. Using a similar derivation that found~\eqref{eqn:cone}, we can model this as the following equality constraint
\begin{equation}
    ({}^w\bm{z} \times {}^g\bm{y})^\top J^\omega_e \dot{\bm{q}} - \epsilon_\phi = c~{}^w\bm{z}^\top {}^g\bm{y},
\end{equation}
where $c \in \R_+$ is a gain to adjust the aggressiveness of the constraint, and $\epsilon_\phi \in \R$ is a slack variable.

\subsection{Objective Weighting} \label{sec:obj-weight}


The objective function is weighted to achieve additional desirable behaviors. The same methods used in~\cite{haviland2021holistic} to prioritize the base DoF when far away from the target and to prioritize the arm DoF when close to the target, as well as to eliminate steady-state error of the end-effector arising due to the presence of the slack variables, are adopted in VMC's objective function. 

Specific to the self-occlusion problem, the most significant conflict in VMC's objectives is between end-effector control and self-occlusion avoidance, particularly for grasping scenarios when the end-effector must eventually reach into the LoS to grasp the target object. This may result in local minima where the QP solution returns zero joint velocities despite not reaching the target. The regulation of the prioritization of these two objectives can be done by constructing the weighting matrix $\Lambda_\epsilon$ with diagonal elements of the form
\begin{equation}
    \lambda(\alpha, \gamma) = \alpha{\lVert \bm{e} \rVert}^\gamma,\label{eqn:weight-func}
\end{equation}
where $\lVert \bm{e} \rVert$ is the translational error between the end-effector and the target, and $\alpha \in \R$ and $\gamma \in \R$ are parameters which adjust how aggressively the weighting grows or decays with this error. By choosing $\gamma>0$ for the diagonal element corresponding to $\epsilon_v$, the self-occlusion avoidance task is prioritized when the robot is far away from the target, and the end-effector control task is prioritized when the robot is close to the target. The local minima issue can be further mitigated by choosing $\gamma<0$ for diagonal elements corresponding to the end-effector angular velocity constraints. This allows the controller greater flexibility in how it can orient the end-effector during its trajectory to avoid self-occlusions, while still servoing the end-effector towards the target goal.

\section{Experiments} \label{sec:experiments}

We evaluate the performance of VMC in three different scenarios (summarized in Table ~\ref{table:scenarios}) to validate the generalizability of the proposed approach, with variations on the robot platform, camera setup, and the number of objects the robot is required to avoid occluding.

\begin{table}[h!]
\setlength{\tabcolsep}{1.5pt}
\small
\begin{tabular*}{\columnwidth}{@{}l@{\extracolsep{\fill}}ccccc@{}}
\toprule
\multicolumn{1}{l}{\textbf{Experiment}} &  \begin{tabular}[c]{@{}c@{}}\textbf{Task(Sim)}\end{tabular} & \begin{tabular}[c]{@{}c@{}}\textbf{Task(Real)}\end{tabular} &  \begin{tabular}[c]{@{}c@{}}\textbf{Robot}\end{tabular} & \begin{tabular}[c]{@{}c@{}}\textbf{Object} \end{tabular} & \begin{tabular}[c]{@{}c@{}}\textbf{\# Objs}\end{tabular} \\ \midrule
Fixed-Base & Reach & Reach & Fixed & Fixed & 5\\
Moving Object & Track & Track & Fixed & Moving & 1\\
Mobile Manip. & Reach & Grasp & Mobile & Fixed & 1 \\
\bottomrule
\end{tabular*}
\caption{Experiment scenarios}
\vspace{-0.15cm}
\label{table:scenarios}
\end{table}

Simulated and real-world experiments are conducted for each scenario. Simulations are run in the Swift Simulator\footnote{\href{https://github.com/jhavl/swift}{https://github.com/jhavl/swift}}, where the robot has perfect knowledge of the world state. For real-world experiments, the robot state is estimated using internal joint sensors, and additionally base odometry for experiments involving a mobile manipulator. Objects are detected using an RGB-D sensor with a simple color segmentation and point cloud clustering algorithm. In real-world experiments, all approaches model the table that the objects are on as a collision object that must be avoided. ROS is used to facilitate the communication between the robot and sensors. All approaches are implemented in Python using the Robotics Toolbox library~\cite{RTBPython}. 

Parameters for VMC are tuned separately for each scenario. The performance of VMC is compared against various reactive and planning baseline controllers. These include:
\begin{itemize}
    \item \textbf{NEO}~\cite{haviland2020neo}: A reactive controller baseline that does not consider occlusion avoidance.
    \item \textbf{MoveIt}: A planning baseline using the default OMPL\footnote{\href{https://github.com/ompl/ompl}{https://github.com/ompl/ompl}} planner that does not consider occlusion avoidance.
    \item \textbf{MoveIt+}: A planning baseline that considers self-occlusion avoidance. This is implemented by modeling LoS as cylindrical collision objects in MoveIt. To improve the plan success rate, a heuristic method is used where the LoS to the grasp object is gradually retracted from the grasp object until a plan is successfully generated.
\end{itemize}

We note that the reactive approaches are automatically at a disadvantage compared to planning methods, due to their single-step horizon, hence should be considered as a separate category. In experiments where the object is static (Sec.~\ref{sec:fixed_base} and Sec.~\ref{sec:mobile_manipulation}), we compare our results to the planning methods only as a reference. For moving object experiments (Sec.~\ref{sec:moving_object}), the planning baselines cannot be used since the tracking of a moving object requires real-time control.



\subsection{Experiment 1: Fixed-Base Manipulation}
\label{sec:fixed_base}

A Franka Emika Panda serial manipulator is tasked with top-down grasping a single specified object in an environment with $4$ other objects that the robot is also required to avoid occluding. The robotic arm always begins in a fixed home position, with a static eye-to-hand camera positioned above the robot to obtain a birds-eye-view of the environment. VMC for this task is implemented by removing the base and camera degrees of freedom from the robotic model introduced in Sec.~\ref{sec:model}. The end-effector orientation is constrained using the method described in Sec.~\ref{sec:vel-control}. For baseline methods, the end-effector grasp orientation is constrained to be top-down, with the gripper perpendicular to the vector between the robotic base and grasp object.


\begin{figure}[t]
\centering
\begin{subfigure}[b]{.195\linewidth}
\includegraphics[width=\linewidth]{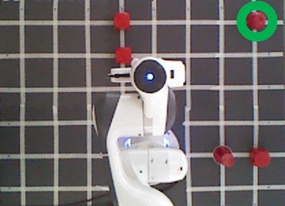}
\end{subfigure}\hfill%
\begin{subfigure}[b]{.195\linewidth}
\includegraphics[width=\linewidth]{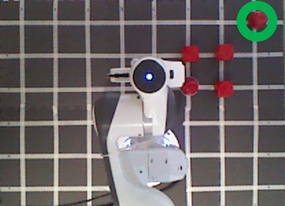}
\end{subfigure}\hfill%
\begin{subfigure}[b]{.195\linewidth}
\includegraphics[width=\linewidth]{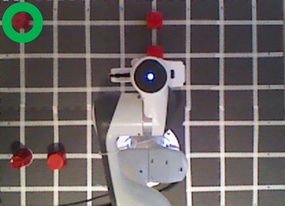}
\end{subfigure}\hfill%
\begin{subfigure}[b]{.195\linewidth}
\includegraphics[width=\linewidth]{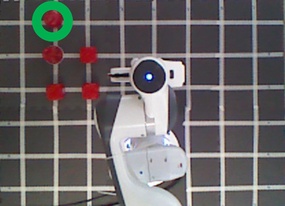}
\end{subfigure}\hfill%
\begin{subfigure}[b]{.195\linewidth}
\includegraphics[width=\linewidth]{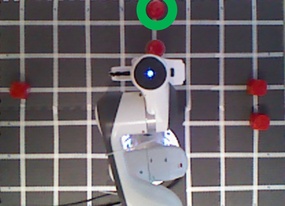}
\end{subfigure}
\\[0.3ex]
\begin{subfigure}[b]{.195\linewidth}
\includegraphics[width=\linewidth]{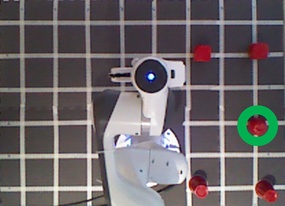}
\end{subfigure}\hfill%
\begin{subfigure}[b]{.195\linewidth}
\includegraphics[width=\linewidth]{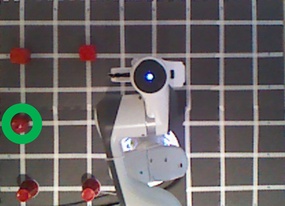}
\end{subfigure}\hfill%
\begin{subfigure}[b]{.195\linewidth}
\includegraphics[width=\linewidth]{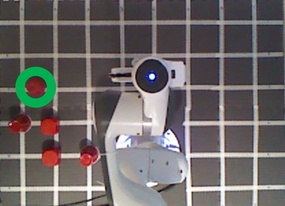}
\end{subfigure}\hfill%
\begin{subfigure}[b]{.195\linewidth}
\includegraphics[width=\linewidth]{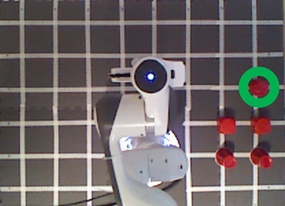}
\end{subfigure}\hfill%
\begin{subfigure}[b]{.195\linewidth}
\includegraphics[width=\linewidth]{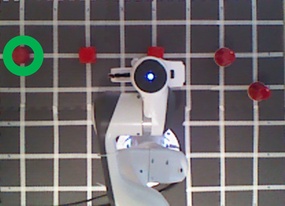}
\end{subfigure}
\caption{The 10 real-world scenarios for fixed-base manipulation. The target object is circled in green.}
\vspace{-0.25cm}
\label{fig:fixed-manipulation-setup}
\end{figure}

\textbf{Simulation experiments:} The object and camera positions are randomized for each scenario. The camera position is randomized within a $\SI{2}{\meter}\times\SI{2}{\meter}\times\SI{0.5}{\meter}$ region above the manipulator. The objects are spawned using random cylindrical coordiantes, at a radial distance between \SI{0.4}{\meter} to \SI{0.7}{\meter} from the robot base, and at a height between \SI{0}{\meter} to \SI{0.5}{\meter} from the robot base.
 Furthermore, the object positions are constrained such that the initial LoS between the camera and the object is not occluded. The controllers are tasked to move directly to the desired pose. 

\textbf{Real-world experiments:} 10 different scenarios shown in Fig.~\ref{fig:fixed-manipulation-setup} are used, with a fixed camera position. A Kinect v1 RGB-D camera is used to track the objects. The controllers are tasked with moving to a pre-grasp pose \SI{10}{\centi\meter} above the object before moving straight down towards the final grasp pose.

\textbf{Metrics}: 
\begin{itemize}
    \item Occlusion (\%): Percentage of frames in which the object is occluded, over all frames for the robot's trajectory, averaged over 5 objects.
    \item Task time (s): Time in seconds elapsed to reach the final desired pose, considered only for successful trials.
    \item Success (\%): The percentage of trials in which the end-effector successfully reached the final desired pose.
\end{itemize}




\subsection{Experiment 2: Moving Object Tracking}
\label{sec:moving_object}

A Franka Emika Panda is tasked with tracking a single moving object by keeping its gripper \SI{30}{\centi\meter} above the object. The same end-effector orientation requirements as Experiment 1 are used.

\begin{figure}[t]
\centering
\pgfplotstableread[col sep=comma]{figures/moving_paths.csv}\data

\begin{tikzpicture}
    \begin{axis}[
            name=mainplot,
            xmin = -0.175, xmax = 0.175,
            ymin = -0.175, ymax = 0.175,
            width = 0.195\textwidth,
            ylabel=\scriptsize$Y$ (m),
            axis equal image,
            xticklabels={,,},
            ticklabel style = {font=\scriptsize},
         ]     
         
        \addplot[] table[x=x1, y=y1]  {\data};
        
    \end{axis}
    \begin{axis}[
            name=secondplot,
            at={(mainplot.north east)},
            xshift=0.0cm,
            anchor=north west,    
            xmin = -0.175, xmax = 0.175,
            ymin = -0.175, ymax = 0.175,
            width = 0.195\textwidth,
            axis equal image,
            xticklabels={,,},
            yticklabels={,,},
         ]     
         
        \addplot[] table[x=x2, y=y2]  {\data};
        
    \end{axis}
    \begin{axis}[
            name=thirdplot,
            at={(secondplot.north east)},
            xshift=0.0cm,
            anchor=north west,      
            xmin = -0.175, xmax = 0.175,
            ymin = -0.175, ymax = 0.175,
            width = 0.195\textwidth,
            axis equal image,
            xticklabels={,,},
            yticklabels={,,},
         ]     
         
        \addplot[] table[x=x3, y=y3]  {\data};
        
    \end{axis}
    \begin{axis}[
            name=fourthplot,
            at={(thirdplot.north east)},
            xshift=0.0cm,
            anchor=north west,      
            xmin = -0.175, xmax = 0.175,
            ymin = -0.175, ymax = 0.175,
            width = 0.195\textwidth,
            axis equal image,
            xticklabels={,,},
            yticklabels={,,},
         ]     
         
        \addplot[] table[x=x4, y=y4]  {\data};
        
    \end{axis}
    \begin{axis}[
            name=fifthplot,
            at={(fourthplot.north east)},
            xshift=0.0cm,
            anchor=north west,      
            xmin = -0.175, xmax = 0.175,
            ymin = -0.175, ymax = 0.175,
            width = 0.195\textwidth,
            axis equal image,
            xticklabels={,,},
            yticklabels={,,},
         ]     
         
        \addplot[] table[x=x5, y=y5]  {\data};
        
    \end{axis}    
    \begin{axis}[
            at={(mainplot.below south west)},
            yshift=-0cm,
            anchor=north west,
            xmin = -0.175, xmax = 0.175,
            ymin = -0.175, ymax = 0.175,
            width = 0.195\textwidth,
            ylabel=\scriptsize$Y$ (m),
            xlabel=\scriptsize$X$ (m),
            axis equal image,
            ticklabel style = {font=\scriptsize}
         ]     
         
        \addplot[] table[x=x6, y=y6]  {\data};
        
    \end{axis}
    \begin{axis}[
            at={(secondplot.below south west)},
            yshift=-0.cm,
            anchor=north west,    
            xmin = -0.175, xmax = 0.175,
            ymin = -0.175, ymax = 0.175,
            width = 0.195\textwidth,
            xlabel=\scriptsize$X$ (m),
            axis equal image,
            yticklabels={,,},
            ticklabel style = {font=\scriptsize}
         ]     
         
        \addplot[] table[x=x7, y=y7]  {\data};
        
    \end{axis}
    \begin{axis}[
            at={(thirdplot.below south west)},
            yshift=-0cm,
            anchor=north west,   
            xmin = -0.175, xmax = 0.175,
            ymin = -0.175, ymax = 0.175,
            width = 0.195\textwidth,
            xlabel=\scriptsize$X$ (m),
            axis equal image,
            yticklabels={,,},
            ticklabel style = {font=\scriptsize},
         ]     
         
        \addplot[] table[x=x8, y=y8]  {\data};
        
    \end{axis}
    \begin{axis}[
            at={(fourthplot.below south west)},
            yshift=-0cm,
            anchor=north west,
            xmin = -0.175, xmax = 0.175,
            ymin = -0.175, ymax = 0.175,
            width = 0.195\textwidth,
            xlabel=\scriptsize$X$ (m),
            axis equal image,
            yticklabels={,,},
            ticklabel style = {font=\scriptsize}
         ]     
         
        \addplot[] table[x=x9, y=y9]  {\data};
        
    \end{axis}
    \begin{axis}[
            at={(fifthplot.below south west)},
            yshift=-0cm,
            anchor=north west,    
            xmin = -0.175, xmax = 0.175,
            ymin = -0.175, ymax = 0.175,
            width = 0.195\textwidth,
            xlabel=\scriptsize$X$ (m),
            axis equal image,
            yticklabels={,,},
            ticklabel style = {font=\scriptsize},
         ]     
         
        \addplot[] table[x=x10, y=y10]  {\data};
        
    \end{axis}
\end{tikzpicture}
\caption{The 10 object trajectories for the moving object tracking experiments.}
\vspace{-0.25cm}
\label{fig:paths}
\end{figure}
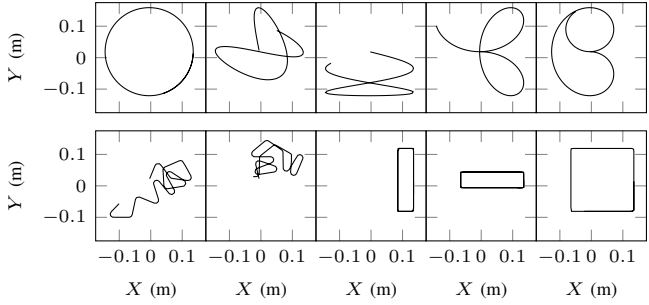

In order to conduct repeatable experiments, the target object is moved using an XY table system shown in Fig.~\ref{fig:intro_fig}, middle row. Similar to Experiment 1, the robot begins in a fixed home position, with a static eye-to-hand camera positioned above the robot. 10 fixed trajectories are created for the object to follow, as shown in Fig.~\ref{fig:paths}. For each trajectory, two experiments are run, where the object moves at \SI{50}{\milli\meter\per\second} and \SI{100}{\milli\meter\per\second}. VMC for this task is implemented similarly to Experiment 1. 

\textbf{Simulation experiments:} The camera positions are randomized (as described in Experiment 1), and a series of affine transformations (scaling, rotation, translation and skew) are applied to the path. All transformation are designed to keep the trajectory within workspace bounds. We assume that the controller can perfectly detect the position and velocity of the object, regardless of whether it is under occlusion.

\textbf{Real-world experiments:} $20$ trials per controller are run (10 trajectories $\times$ 2 speeds). Whenever the camera loses sight of the object due to occlusions, we assume it is positioned at the last observed location.

\textbf{Metrics}: 
\begin{itemize}
    \item Occlusion (\%): Percentage of frames in which the object is occluded, over all frames for the robot's trajectory.
    \item Perception Error (mm): The mean $xy$-distance between the measured position of the object and the ground-truth position. Only defined for real-world experiments since we assume perfect perception in simulation.
    \item Tracking error (mm): The mean $xy$-distance between the end-effector position and the object position
\end{itemize}



\subsection{Experiment 3: Mobile Manipulation}
\label{sec:mobile_manipulation}

A Fetch Mobile Manipulator, which consists of a non-holonomic mobile base and a Primesense Carmine RGB-D camera mounted within the 2 DoF head, is tasked with grasping an object (real) or reaching to it (sim). The robot's base begins at a distance away from the target object, and does not necessarily begin pointing towards the object. The Fetch must therefore utilize both its base and arm together to navigate towards, and grasp or reach, the object.

VMC for this task is implemented using the full robot model described in Sec.~\ref{sec:model}. The end-effector is tasked to reach a specific end-effector orientation. The NEO baseline is similarly modified for a mobile manipulator platform~\cite{haviland2021holistic}. Two methods for camera control are additionally applied for NEO, and treated as separate baselines. The first method is to incorporate coordinated camera control (CCC) using the method in Sec.~\ref{sec:fov}. The second method is to incorporate independent camera control (ICC) using a simple proportional controller to control the camera head joints. The MoveIt baselines are implemented by sequentially rotating the base towards the object, driving the base forwards, then using MoveIt to plan a grasp trajectory once the base has stopped. The same ICC method is used throughout the robot's trajectory.

\textbf{Simulation Experiments:} The robot is tasked with reaching a desired pose. As contact physics is not supported in the Swift simulator, grasping performance is not evaluated. Both the position and orientation of the desired pose are randomized within a $\SI{5}{\meter}\times\SI{5}{\meter}\times\SI{0.5}{\meter}$ region around the robot. The initial joint configuration is set to a random valid configuration that does not self-collide. 

\textbf{Real-world experiments:} The robot is tasked with grasping an object. For all controllers the orientation of the end-effector is constrained to perform a side grasp (parallel to the ground) with the gripper pointing in the direction between the initial robot position and target. We considered a combination of 3 initial arm configurations and 5 initial base positions for a total of 15 trials per controller. To account for initial base positions pointing away from the object, an initial guess of the object position is provided, which is then corrected once the object is in view of the robot. The grasping is performed by first moving the end-effector to a waypoint \SI{10}{\centi\meter} away from the object in the ${}^g\bm{x}$ direction, before reaching straight forwards to perform a side-grasp on the object. The grippers are closed to perform a grasp once the end-effector has reached its goal. 

\textbf{Metrics}: 
\begin{itemize}
    \item Occlusion (\%): Percentage of frames in which the object is occluded, over all frames for the robot's trajectory. This includes FoV occlusions in addition to self-occlusions.
    \item Task time (s): Time elapsed to reach the final desired pose. Defined only for successful trials.
    \item Success (\%) (Sim): The percentage of trials in which end-effector successfully reached the final desired pose.
    \item Success (\%) (Real): The percentage of trials in which the robot successfully grasped the object.
\end{itemize}


\section{Results} \label{sec:results}

\subsection{Experiment 1: Fixed-Base Manipulation}

\begin{table}[t]
\small
\begin{tabular*}{\columnwidth}{@{}l@{\extracolsep{\fill}}ccc@{}}
\toprule
\multicolumn{1}{c}{} & \textbf{\begin{tabular}[c]{@{}c@{}}Occlusion (\%)\end{tabular}} & \textbf{\begin{tabular}[c]{@{}c@{}}Task time (s)\end{tabular}} & \textbf{\begin{tabular}[c]{@{}c@{}}Success (\%)\end{tabular}} \\ \midrule
\textbf{VMC (Ours)} & \textbf{4.1} & 6.6 & 94.5 \\
\textbf{NEO}~\cite{haviland2020neo} & 9.9 & \textbf{6.4} & \textbf{99.2} \\ \hdashline\noalign{\vskip 0.5ex}
\textbf{MoveIt} & 9.6 & \textbf{9.6} & \textbf{100.0} \\
\textbf{MoveIt+} & \textbf{1.7} & 16.6 & 72.6 \\ \bottomrule
\end{tabular*}
\caption{Simulation results for fixed-base manipulation (1000 trials). Reactive and planning approaches are separated with the dashed line.}
\label{table:static-sim}
\end{table}

\begin{table}[t]
\small
\begin{tabular*}{\columnwidth}{@{}l@{\extracolsep{\fill}}ccc@{}}
\toprule
\multicolumn{1}{c}{} & \textbf{\begin{tabular}[c]{@{}c@{}}Occlusion (\%)\end{tabular}} & \textbf{\begin{tabular}[c]{@{}c@{}}Task time (s)\end{tabular}} & \textbf{\begin{tabular}[c]{@{}c@{}}Success (\%)\end{tabular}} \\ \midrule
\textbf{VMC (Ours)} & \textbf{19.4} & 17.2 & \textbf{100} \\
\textbf{NEO}~\cite{haviland2020neo} & 45.9 & \textbf{14.9} & \textbf{100}\\ \hdashline\noalign{\vskip 0.5ex}
\textbf{MoveIt} & 35.7 & \textbf{12.4} & \textbf{100} \\
\textbf{MoveIt+} & \textbf{5.1} & 38.7 & 80 \\ \bottomrule
\end{tabular*}
\caption{Real-world results for fixed-base manipulation (10 trials). Reactive and planning approaches are separated with the dashed line.}
\vspace{-0.25cm}
\label{table:static-real}
\end{table}

Simulation and real-world experiment results are shown in Table~\ref{table:static-sim} and Table~\ref{table:static-real}, respectively. When we compare the performance of reactive methods, for both simulated and real-world experiments, we see that the baseline NEO method resulted in almost $2.5$ times more occlusions than VMC. The task time for VMC was slightly higher, which was expected because the robot does more maneuvering to avoid occluding the objects. On the other hand, VMC suffered from a $4.7\%$ drop in success compared to the NEO baseline in simulated experiments, likely due to the additional constraints resulting in a higher chance of the optimizer falling into local minima.

When we compare the reactive methods to planning based methods, the MoveIt+ baseline scored the lowest occlusion rates by far. However, the success rate was $72.6\%$ in simulation and $80\%$ in real experiments, which is much lower than other methods. This may indicate that the MoveIt+ baseline is passing on more difficult cases. Moreover, MoveIt+ generates inefficient paths that take $2$ to $3$ times longer to execute compared to other methods. The default planning method MoveIt was successful in each trial, however the occlusion rate was much higher than VMC as expected ($19.4\%$ vs $35.7\%$).

The fixed-base experiments show that while VMC marked about a $2.5$ fold decrease in occlusions compared to the reactive controller baseline, it came at the expense of slightly lower success and increased task time. Compared to the planning baselines, VMC offers a trade-off by keeping the occlusion of the objects relatively low while keeping the success rate high.

\subsection{Experiment 2: Moving Object Tracking}

\begin{table}[t]
\small
\begin{tabular*}{\columnwidth}{@{}l@{\extracolsep{\fill}}cc@{}}
\toprule
\multicolumn{1}{c}{} & \textbf{\begin{tabular}[c]{@{}c@{}}Occlusion (\%)\end{tabular}} & \textbf{\begin{tabular}[c]{@{}c@{}}Tracking error (mm)\end{tabular}} \\ \midrule
\textbf{VMC (Ours)} & \textbf{3.2} & 188.0 \\
\textbf{NEO}~\cite{haviland2020neo} & 43.0 & \textbf{150.9} \\ \bottomrule
\end{tabular*}
\caption{Simulated results for moving object tracking (1000 trials)}
\label{table:moving-sim}
\end{table}

\begin{table}[t]
\small
\begin{tabular*}{\columnwidth}{@{}l@{\extracolsep{\fill}}ccc@{}}
\toprule
\multicolumn{1}{c}{} & \textbf{\begin{tabular}[c]{@{}c@{}}Occlusion (\%)\end{tabular}} & \textbf{\begin{tabular}[c]{@{}c@{}}Perception\\ error (mm)\end{tabular}} & \textbf{\begin{tabular}[c]{@{}c@{}}Tracking\\ error (mm)\end{tabular}} \\ \midrule
\textbf{VMC (Ours)} & \textbf{0.1} & \textbf{7.9} & 108.9 \\
\textbf{NEO}~\cite{haviland2020neo} & 80.6 & 49.4 & \textbf{70.7} \\ \bottomrule
\end{tabular*}
\caption{Real-world results for moving object tracking (20 trials)}
\vspace{-0.2cm}
\label{table:moving-real}
\end{table}

Simulation and real-world experiment results are shown in Table~\ref{table:moving-sim} and Table~\ref{table:moving-real}, respectively. VMC resulted in much less occlusions, especially for real robot experiments. Particularly from a top-down camera perspective, NEO tends to occlude the object whenever it moves towards it, and new observations can only be made when the object moves outside of the occlusion area. In contrast, VMC was much more successful in avoiding self-occlusions. Brief failure modes for VMC occur when the object makes sudden changes on its trajectory such as on high curvature paths. This is not surprising since our approach does not attempt to predict the future position of the object.

As a result of VMC's superior self-occlusion avoidance, its perception error in real-world experiments is predominantly made up of measurement noise, as opposed to NEO which suffers due to its large occlusion rate. As expected, VMC has a higher tracking error due to its self-occlusion avoidance requirement, which often prevents the robot from reaching the target pose. However, the difference in tracking error between VMC and NEO is within the \SI{5}{\centi\meter} minimum distance restriction ($d_s$ in Eq. \ref{eqn:vel-damper}) imposed between the end-effector and the LoS for VMC to avoid self-occlusions.


\subsection{Experiment 3: Mobile Manipulation}

Simulation and real-world experiment results are shown in Table~\ref{table:mobile-sim} and Table~\ref{table:mobile-real}, respectively. When we compare the performance of reactive methods, VMC suffered from the least amount of occlusions for both real and simulated experiments, with about a $20$ percentile points difference. The gain in lower occlusion rates accompanied a slightly longer task time and lower success rate. This finding is consistent with the results from Experiment 1. In simulation, NEO~+~CCC baseline had the least task time and highest success rate, whereas in real experiments the improved visibility of VMC led to vastly improved grasp success rates in the real-world experiments, which validates the motivation of avoiding occlusions for closed-loop visual servoing. Both simulated and real experiments successfully validate that VMC is able to improve the visibility of the target object in grasping task with a mobile base without sacrificing too much from other task metrics.

When we compare the reactive methods to planning based methods, the MoveIt baseline suffered from a lower grasp success rate due to the open-loop nature of the controllers, as errors in the measured position of the object could not be corrected during the course of the arm movement. Similar to Experiment 1, MoveIt+ baseline had the lowest occlusion rate among all methods, however it came at the expense of inefficient paths (as indicated by increased task times) and relatively low task success.

A couple of emergent behaviors were observed when we employed VMC for mobile manipulation. A common behavior was for the base to take a curved path towards the target so that it could obtain a side-on view of the target which is less susceptible to self-occlusions. This behavior is shown in Fig.~\ref{fig:intro_fig}, bottom row. Another common behavior was for the controller to lift the torso up to obtain a higher birds-eye-view-like perspective of the scene which offers a less obstructed view.








\begin{table}[t]
\small
\begin{tabular*}{\columnwidth}{@{}l@{\extracolsep{\fill}}ccc@{}}
\toprule
\multicolumn{1}{c}{} & \textbf{\begin{tabular}[c]{@{}c@{}}Occlusion (\%)\end{tabular}} & \textbf{\begin{tabular}[c]{@{}c@{}}Task time (s)\end{tabular}} & \textbf{\begin{tabular}[c]{@{}c@{}}Success (\%)\end{tabular}} \\ \midrule
\textbf{VMC (Ours)} & \textbf{27.0} & 8.4 & 97.3 \\
\textbf{NEO}~\cite{haviland2021holistic} \textbf{+ CCC} & 38.2 & \textbf{7.6} & \textbf{99.5} \\
\textbf{NEO}~\cite{haviland2021holistic} \textbf{+ ICC} & 48.0 & 10.8 & 98.9 \\ \hdashline\noalign{\vskip 0.5ex}
\textbf{MoveIt} & 26.9 & \textbf{13.8} & \textbf{92.8} \\
\textbf{MoveIt+} & \textbf{22.0} & 15.0 & 87.5 \\ \bottomrule
\end{tabular*}
\caption{Simulation results for mobile manipulation (1000 trials). Reactive and planning approaches are separated with the dashed line.}
\label{table:mobile-sim}
\end{table}

\begin{table}[t]
\small
\begin{tabular*}{\columnwidth}{@{}l@{\extracolsep{\fill}}ccc@{}}
\toprule
\multicolumn{1}{c}{} & \textbf{\begin{tabular}[c]{@{}c@{}}Occlusion (\%)\end{tabular}} & \textbf{\begin{tabular}[c]{@{}c@{}}Task time (s)\end{tabular}} & \textbf{\begin{tabular}[c]{@{}c@{}} Success (\%)\end{tabular}} \\ \midrule
\textbf{VMC (Ours)} & \textbf{33.7} & \textbf{57.0} & \textbf{80.0} \\
\textbf{NEO}~\cite{haviland2021holistic} \textbf{+ CCC} & 51.8 & 63.5 & 33.3 \\
\textbf{NEO}~\cite{haviland2021holistic} \textbf{+ ICC} & 58.4 & 59.6 & 46.7 \\ \hdashline\noalign{\vskip 0.5ex}
\textbf{MoveIt} & 40.0 & \textbf{62.7} & \textbf{73.1} \\
\textbf{MoveIt+} & \textbf{32.7} & 73.6 & 66.7 \\ \bottomrule
\end{tabular*}
\caption{Real-world results for mobile manipulation (15 trials). Reactive and planning approaches are separated with the dashed line.}
\vspace{-0.2cm}
\label{table:mobile-real}
\end{table}



\section{Conclusion} \label{sec:conclusion}

We present the Visibility Maximization Controller, a reactive controller to perform self-occlusion avoidance that can be applied to a variety of robotic platforms and scenarios. We propose an approach based on quadratic programming that includes a soft constraint to keep the line of sight between the object(s) and the camera clear, and solves for optimal and feasible joint velocities which holistically control the robot and camera together to prevent the arm from blocking the view of the object.

The performance of the proposed controller is validated in a range of randomized simulation experiments and real-world experiments which include static and mobile bases, static and controlled camera configurations, scenarios with single or multiple, and static or moving objects. These experiments show that the proposed controller successfully reduces self-occlusion rates while remaining robust to complex environments and not sacrificing significantly on task efficiency. Real-world experiments with closed-loop mobile grasping demonstrate the importance of retaining vision of the object to improve the chance of grasp success. The baseline planning-based methods achieved a lower occlusion rate than the proposed reactive controller, at the expense of significantly lowered task success and efficiency. Overall, the results showed that a there is a trade-off between the occlusion rates and task efficiency. Future work includes more exploration into this trade-off. Controllers with multi-step horizons or planning-based methods can be useful in generating globally optimal trajectories. Predicting the future motion of the moving objects can further improve the performance. The approach can also be extended to occlusions caused by external obstacles that are interrupting the line of sight, which would result in a more complete occlusion-avoidance control method.




\bibliographystyle{IEEEtran}
\bibliography{refs}

\begin{thebibliography}{10}
\providecommand{\url}[1]{#1}
\csname url@samestyle\endcsname
\providecommand{\newblock}{\relax}
\providecommand{\bibinfo}[2]{#2}
\providecommand{\BIBentrySTDinterwordspacing}{\spaceskip=0pt\relax}
\providecommand{\BIBentryALTinterwordstretchfactor}{4}
\providecommand{\BIBentryALTinterwordspacing}{\spaceskip=\fontdimen2\font plus
\BIBentryALTinterwordstretchfactor\fontdimen3\font minus
  \fontdimen4\font\relax}
\providecommand{\BIBforeignlanguage}[2]{{%
\expandafter\ifx\csname l@#1\endcsname\relax
\typeout{** WARNING: IEEEtran.bst: No hyphenation pattern has been}%
\typeout{** loaded for the language `#1'. Using the pattern for}%
\typeout{** the default language instead.}%
\else
\language=\csname l@#1\endcsname
\fi
#2}}
\providecommand{\BIBdecl}{\relax}
\BIBdecl

\bibitem{corke1994high}
P.~I. Corke, ``High-performance visual closed-loop robot control,'' Ph.D.
  dissertation, 1994.

\bibitem{lenz2015deep}
I.~Lenz, H.~Lee, and A.~Saxena, ``Deep learning for detecting robotic grasps,''
  \emph{The International Journal of Robotics Research}, 2015.

\bibitem{mahler2017dex}
J.~Mahler \emph{et~al.}, ``Dex-net 2.0: Deep learning to plan robust grasps
  with synthetic point clouds and analytic grasp metrics,'' in \emph{RSS},
  2017.

\bibitem{hutchinson1996tutorial}
S.~Hutchinson, G.~D. Hager, and P.~I. Corke, ``A tutorial on visual servo
  control,'' \emph{IEEE Transactions on Robotics and Automation}, 1996.

\bibitem{haviland2020control}
J.~Haviland, F.~Dayoub, and P.~Corke, ``Control of the final-phase of
  closed-loop visual grasping using image-based visual servoing,'' \emph{arXiv
  preprint arXiv:2001.05650}, 2020.

\bibitem{flandin2000eye}
G.~Flandin, F.~Chaumette, and E.~Marchand, ``Eye-in-hand/eye-to-hand
  cooperation for visual servoing,'' in \emph{ICRA}, 2000.

\bibitem{haviland2021holistic}
J.~{Haviland}, N.~{Sünderhauf}, and P.~{Corke}, ``A holistic approach to
  reactive mobile manipulation,'' \emph{IEEE Robotics and Automation Letters},
  vol.~7, no.~2, pp. 3122--3129, 2022.

\bibitem{sciavicco1988solution}
L.~Sciavicco \emph{et~al.}, ``A solution algorithm to the inverse kinematic
  problem for redundant manipulators,'' \emph{J. on Robotics and Auto.}, 1988.

\bibitem{haviland2020purelyreactive}
J.~Haviland and P.~Corke, ``A purely-reactive manipulability-maximising motion
  controller,'' \emph{arXiv preprint 2002.11901}, 2020.

\bibitem{haviland2020neo}
J.~{Haviland} and P.~{Corke}, ``{NEO}: A novel expeditious optimisation
  algorithm for reactive motion control of manipulators,'' \emph{IEEE Robotics
  and Automation Letters}, vol.~6, no.~2, pp. 1043--1050, 2021.

\bibitem{mansard2009unified}
N.~Mansard, O.~Khatib, and A.~Kheddar, ``A unified approach to integrate
  unilateral constraints in the stack of tasks,'' \emph{IEEE T-RO}, 2009.

\bibitem{khatib1986real}
O.~Khatib, ``Real-time obstacle avoidance for manipulators and mobile robots,''
  in \emph{Autonomous robot vehicles}.\hskip 1em plus 0.5em minus 0.4em\relax
  Springer, 1986.

\bibitem{baumann2010path}
M.~Baumann, S.~Leonard, E.~A. Croft, and J.~J. Little, ``Path planning for
  improved visibility using a probabilistic road map,'' \emph{T-RO}, 2010.

\bibitem{marchand1998dynamic}
E.~Marchand and G.~D. Hager, ``Dynamic sensor planning in visual servoing,'' in
  \emph{ICRA}, 1998.

\bibitem{nicolis2018occlusion}
D.~Nicolis, M.~Palumbo, A.~M. Zanchettin, and P.~Rocco, ``Occlusion-free visual
  servoing for the shared autonomy teleoperation of dual-arm robots,''
  \emph{IEEE Robotics and Automation Letters}, 2018.

\bibitem{tarabanis1996computing}
K.~Tarabanis, R.~Y. Tsai, and A.~Kaul, ``Computing occlusion-free viewpoints,''
  \emph{IEEE Trans. on Pattern Analysis and Mach. Intel.}, 1996.

\bibitem{sugiura2007real}
H.~Sugiura, M.~Gienger, H.~Janssen, and C.~Goerick, ``Real-time collision
  avoidance with whole body motion control for humanoid robots,'' in
  \emph{IROS}, 2007.

\bibitem{suzuki2020coordinated}
T.~Suzuki and K.~Sekiyama, ``A coordinated control system for dual-arm mobile
  manipulator balancing grasping and viewpoint selection,'' in \emph{IEEE/SICE
  International Symposium on System Integration}, 2020.

\bibitem{logothetis2018model}
M.~Logothetis, G.~C. Karras, S.~Heshmati-Alamdari, P.~Vlantis, and K.~J.
  Kyriakopoulos, ``A model predictive control approach for vision-based object
  grasping via mobile manipulator,'' in \emph{IROS}, 2018.

\bibitem{grandia2019feedback}
R.~Grandia, F.~Farshidian, R.~Ranftl, and M.~Hutter, ``Feedback mpc for
  torque-controlled legged robots,'' in \emph{IEEE/RSJ International Conference
  on Intelligent Robots and Systems (IROS)}, 2019.

\bibitem{haviland2020systematic}
J.~Haviland and P.~Corke, ``A systematic approach to computing the manipulator
  jacobian and hessian using the elementary transform sequence,'' \emph{arXiv
  preprint arXiv:2010.08696}, 2020.

\bibitem{faverjon1987local}
B.~Faverjon and P.~Tournassoud, ``A local based approach for path planning of
  manipulators with a high number of degrees of freedom,'' in \emph{IEEE
  International Conference on Robotics and Automation}, 1987.

\bibitem{RTBPython}
P.~Corke and J.~Haviland, ``Not your grandmother’s toolbox – the robotics
  toolbox reinvented for {P}ython,'' in \emph{IEEE International Conference on
  Robotics and Automation (ICRA)}, 2021.

\end{thebibliography}

\end{document}